\documentclass[runningheads]{llncs}
\usepackage{graphicx}
\usepackage[english]{babel}
\usepackage[utf8]{inputenc}
\usepackage{algorithm}
\usepackage[noend]{algpseudocode}
\usepackage{diagbox}
\usepackage{color}
\usepackage{url} 
\usepackage{breakurl} 
\usepackage{hyperref} 
\usepackage{amsmath}

\begin{document}
\title{A Semi-Supervised Framework for Misinformation Detection}
%
%
\author{Yueyang Liu \and
Zois Boukouvalas \and
Nathalie Japkowicz}
\authorrunning{Y. Liu et al.}
%
\institute{American University, Washington, D.C. 20016, USA \\
\email{yueyang.liu@student.american.edu \\ \{boukouva,japkowic\}@american.edu}}
\maketitle              
\begin{abstract}
The spread of misinformation in social media outlets has become a prevalent societal problem and is the cause of many kinds of social unrest. Curtailing its prevalence is of great importance and machine learning has shown significant promise. However, there are two main challenges when applying machine learning to this problem. First, while much too prevalent in one respect, misinformation, actually, represents only a minor proportion of all the postings seen on social media. Second, labeling the massive amount of data necessary to train a useful classifier becomes impractical. Considering these challenges, we propose a simple semi-supervised learning framework in order to deal with extreme class imbalances that has the advantage, over other approaches, of using actual rather than simulated data to inflate the minority class. 
We tested our framework on two sets of Covid-related Twitter data and obtained significant improvement
in F1-measure on extremely imbalanced scenarios, as compared to simple classical and deep-learning data generation methods such as SMOTE, ADASYN, or GAN-based data generation.

\end{abstract}
\section{Introduction}

The spread of misinformation in social media outlets has become a prevalent societal problem and is the cause of many kinds of social unrest. Curtailing its prevalence is of great importance {and machine learning advances have shown significant promise for the detection of misinformation \cite{Islam2020DeepLF}. However, 
to build a reliable model 
a large data set of reliable posts as well as posts containing misinformation is needed. In practice, this is not feasible since detecting posts containing misinformation is inherently a class imbalanced problem: the majority of posts are reliable whereas a very small minority contains misinformation. For instance,
according to The Verge, an American technology news website operated by Vox Media, Twitter removed 2,230 misleading tweets between March 16 and April 18, 2020\footnote{https://www.theverge.com/2020/4/22/21231956/twitter-remove-covid-19-tweets-call-to-action-harm-5g}. Given that, on average, 6,000 tweets are tweeted every second\footnote{https://www.internetlivestats.com/twitter-statistics/}, 
the class imbalance ratio is around 0.000014\% for that month, or 1 unreliable Tweet for every 71,428 reliable ones, an extreme imbalance ratio.}

The class imbalance problem has been pervasive in the Machine Learning field for over two decades \cite{Japkowicz2002TheCI,he2009learning,branco2016survey,krawczyk2016learning,johnson2019survey}. The class imbalance problem and issues related to it are, in part, responsible for questions of algorithmic bias and fairness that are very much on researcher's and the public's mind now that machine learning algorithms are routinely deployed in applications that directly affect people. Over the years, many techniques for dealing with class imbalances have been proposed including classical methods for inflating the minority class such as SMOTE \cite{chawla_smote_2002} and ADASYN \cite{He2008ADASYNAS} 
and Deep-Learning based methods such as 
DEAGO \cite{Bellinger2017ManifoldbasedSO} and 
GAMO \cite{Mullick2019GenerativeAM}, which use an autoencoder and  a Generative Adversarial Network, respectively.  
One of the issues with previously proposed minority-class oversampling methods for the class imbalance problem is that either the data used to inflate the minority class is real but simply repeated from the existing minority class, as in random oversampling \cite{Japkowicz2002TheCI}, or it is artificial as in SMOTE \cite{chawla_smote_2002}. Random oversampling is not an acceptable solution given that it is known to cause overfitting \cite{drummond2003c4}. Artificial oversampling, while not overfitting as much as random oversampling, generates artificial data. While this kind of data approximates real data fairly well in continuous domains such as computer vision, it is not as representative in non-continuous domains such as text \cite{Hu2017TowardCG}. This is the reason why, instead of proposing a text generation method to inflate the minority class, this paper proposes a semi-supervised method which, instead of generating new text artificially, relies on the available unlabeled data. A deep-learning method is used to label the data, but not to generate new text.

Semi-Supervised Learning for text data is not new and was first proposed in the context of class imbalance in \cite{Li2011SemiSupervisedLF}. However, while the class imbalance was present in that study, it was not as extreme as it is in the case of misinformation detection since the authors use an undersampling of the majority class strategy to bring the size of the two classes closer to one another. In our case, we are dealing with such an extremely imbalanced data set that solutions of the type proposed in \cite{Li2011SemiSupervisedLF} would not apply.
Semi-supervised learning in class-imbalanced setting is also not new. Authors in \cite{DBLP:journals/corr/abs-2002-06815} review existing approaches and propose their own. However, they focus on algorithmic modifications rather than the simpler and more practical re-sampling strategy.

Our framework is similar to standard approaches previously designed to tackle the class imbalance problem, but it differs from them in one important way. On the one hand, like methods such as SMOTE, 
GAMO and so on, it proposes to oversample the minority class, but on the other hand,
unlike these approaches, instead of using generated samples 
it identifies candidates from the unlabeled data set to inflate the minority class with.
Although the search for such candidates could be extremely costly, we show how the use of a K-D Tree makes it tractable.


We evaluate our framework on two data sets related to Covid-19 misinformation in social media, the one collected and curated in-house, early in the pandemic  \cite{boukouvalas_independent_2020}, and a data set obtained from English COVID-19 Fake News and Hindi Hostile Posts data set\cite{patwa2021overview}. Our framework takes two forms: the direct approach in which the labeled minority samples alone are used to search the unlabeled data set; and the indirect approach, designed to increase the diversity of the search,  where artificial data are first generated from the minority class and these samples, along with the original minority samples, are used to search the unlabeled set. Different instantiations of these approaches are compared to traditional ways of overcoming the class imbalance problem and to the results obtained on the original imbalanced data set. 
The results show that the direct implementation of our framework is superior to the indirect approach, which in turn, is superior to the traditional approaches. All of them improve upon not attempting to counter the class imbalance problem.

{The remainder of the paper is organized as follows. In section 2, we discuss previous work on oversampling methods for class imbalances, semi-supervised learning, and discuss the functionality of K-D Trees.  Section~3 introduces our framework and discusses its direct and indirect instantiations. The experimental set-up is discussed in Section 4, and the results of our experiments are presented  in Section 5.  Section 6 concludes the paper.}

\section{Related work}
This section reviews previous work related to this study. We first discuss the methods for inflating the minority class that were previously proposed in the context of the class imbalance problem, and we then move to a discussion of previous work in semi-supervised learning, especially for class imbalanced data. We then describe the K-D Tree data structure along with the Nearest Neighbor Search algorithm associated with it, and  used in this paper. 

\subsection{The class imbalance problem}
The class imbalance problem corresponds to the problem where one or more classes are represented by a much smaller proportion of examples than the other classes. In such cases, classifiers tend to ignore the data from the minority class causing systematic misclassification of these classes. The problem has been well documented for a number of years \cite{Japkowicz2002TheCI,he2009learning,branco2016survey,krawczyk2016learning,johnson2019survey}.
It is typically addressed in one of four ways: undersampling, oversampling, re-weighting the classes, and one-class classification. In this study, we focus on oversampling, which was shown, over the years, to be a reliable and simple approach to deal with the class imbalance problem. As discussed in  \cite{drummond2003c4}, random oversampling is not effective as it causes overfitting of the minority class instances. Instead, it is important to generate instances that are closely related to the original instances, but not exact replicas. We review the three approaches used here to re-balance the minority class: SMOTE, ADASYN, and a Generative Adversarial Network (GAN) combined with a Variational Autoencoder (VAE).\footnote{A DC-GAN and a VAE were also tried separately, but since the VAE-GAN obtained the best results, it is the only approach from the generative series of experiments reported in this work.}
\subsubsection{ SMOTE and ADASYN } 
The Synthetic Minority Oversampling Technique (SMOTE) \cite{chawla_smote_2002}, is an oversampling approach that generates minority class instances to balance data sets. It searches for the $K$ closest minority neighbors of each sample point in the minority class using the Euclidean distance. For each minority class sample $x_i$ , the algorithm randomly chooses a number of samples from its $K$ closet minority neighbors denoted as $x_i(nn)$. For each $x_i$, we generate new samples using the following formula
$$x_i^{new}=x_i+\alpha\left(x_i(nn)-x_i\ \right),$$
where $\alpha$ is a random number from 0 to 1. For the purpose of this work, we use the implementation found in the imbalanced-learn python library, with $K=2$.

Adaptive Synthetic Sampling (ADASYN) is an other oversampling method \cite{He2008ADASYNAS}, which, instead of synthesizing the same number of samples for each minority sample like SMOTE, it uses a mechanism to automatically determine how many synthetic samples need to be generated for each minority sample. For each minority class sample $x_i$, with its $K$ nearest neighbors $x_i(nn)$, it is possible to calculate the ratio $r_i=\frac{x_i(nn)}{K}$, and then normalize this ratio to obtain the density distribution  $\Gamma_{i}=\frac{r_{i}}{\sum r_i}$. The calculation of a synthetic sample is obtained by $g_i=\Gamma_{i} \times G$, where $G$ is the discrepancy between 2 classes. For the purpose of this work, we use the ADASYN package from the imbalanced-learn python library, with $K=2$.

\subsubsection{Generating Adversarial Networks (GANs)} 
A generative adversarial network (GAN) \cite{goodfellow2014generative} consists of two neural networks: a generator $G$ and a discriminator $D$. These two networks are trained in opposition to one another. The generator $G$ takes as input a random noise vector $z \sim p(z)$ and outputs $m$ sample  ${\widetilde{x}}^i=G\left(z^i\right)$. The discriminator $D$ receives as input the training sample $x^i$ and ${\widetilde{x}}^i$  and uses the loss function $$ \check{V}_{max}=\frac{1}{m} \sum_{i=1}^{m} \log D\left(x^{i}\right)+\frac{1}{m} \sum_{i=1}^{m} \log \left(1-D\left(\check{x}^{i}\right)\right)
$$ to update the discriminator $D$ ’s parameters $\theta_d$;\\ Then it uses another random noise vector $z \sim p(z)$ and loss function: $$
\check{V}_{min}=\frac{1}{m} \sum_{i=1}^{m} \log \left(1-\left(D\left(G\left(z^{i}\right)\right)\right)\right.
$$
to update the generator $G$’s parameters $\theta_g$


A VAE-GAN is a Variational Autoencoder combined with a Generative Adversarial Network \cite{larsen_autoencoding_2016}. It uses a GAN discriminator that can be used in place of a Variational Autoencoder (VAE) decoder to learn the loss function. The VAE loss function equals the negative sum of the expected log-likelihood (the reconstruction error) and a prior regularization term as well as a binary cross-entropy in the discriminator. This is what was used in this work.

\subsection{Semi-supervised learning } 
Semi-supervised learning is a learning paradigm in which unlabeled data are leveraged along with the labeled training set to help improve classification performance 
\cite{van2020survey}. Semi-supervised learning is highly practical since labeling work is usually costly in terms of manpower and material resources \cite{zhu2005semi}. There are two common methods used in semi-supervised learning \cite{zhou_2020}. The first one relies on the ``clustering assumption" which assumes that the data follows a cluster structure and that samples in the same cluster belong to the same category. This is the approach we follow in our research, as will be discussed in the next section. Another method follows the ``manifold assumption" which assumes that the data is distributed on a manifold structure and that adjacent samples on that structure should output similar values. In such methods, the degree of proximity is often used to described the degree of similarity. The manifold hypothesis can be viewed as a generalization of the clustering hypothesis. Therefore,  with no restriction on the format of the output value, the manifold assumption is more widely applicable than the clustering assumption as it can be used for a variety of learning tasks. Since we are working in the context of detection, a special case of classification, the ``clustering assumption" is sufficient for our purposes.

As discussed in the introduction, several works have looked at the question of semi-supervised learning in the context of the class imbalance problem  \cite{DBLP:journals/corr/abs-2002-06815,Li2011SemiSupervisedLF,Yang2020RethinkingTV}. While interesting, these works are not closely related to the work in this paper since they do not consider the approach that consists of inflating the minority class nor do they look at the extremely imbalanced context. 


\subsection{K-Dimensional tree and nearest neighbor search}
The search for nearest neighbors that we propose to undertake to identify data close to the labeled minority class data is computationally expensive.  K-D Trees (K-dimension trees) are a kind of binary trees, which divide the k-dimensional data space hierarchically, and stores the points in the k-dimension space in order to query its tree-shaped data structure afterwards \cite{otair2013approximate}. Using K-D Trees can reduce the search space compared to other clustering algorithm such as K-Nearest Neighbors which have a time complexity of $O(n \times m)$, where $n$ is the size of the data set and $m$ is its dimension. Since our corpus embedding method generate a high dimensional corpus feature matrix$(n \times 1024)$, to reduce the search time complexity, we used K-D Tree search rather than other clustering algorithms. Commonly, the K-D Tree can be constructed in $O(n \log{n})$, and the query algorithm has a running time $O(\sqrt{n}+k)$ where $k$ is the number of nearest points reported.


\section{Our Framework} 
We propose a data augmentation method which, instead of randomly sampling from the minority class or generating synthetic minority samples based on the existing minority samples, leverages the unlabeled data set.
The method is geared at non-continuous feature spaces such as those emanating from text applications, which present particular difficulty for data generation processes. 

Our approach works on binary data and takes as input a labeled imbalanced data set $LI$ and an unlabeled data set $U$, drawn from the same population. It outputs a labeled balanced data set $LB$ that is then used for classification. It
works as follows:
\begin{description}
\item[Step 1:] Pre-process the $LI$ and $U$ using the same embedding process and separate the majority from the minority samples 
\item[Step 2 (optional):] Use the minority set as a sample to generate synthetic data resembling that data.
\item[Step 3:] Construct a K-D Tree from the minority samples of Step 1 or the augmented minority samples from Step 2.
\item[Step 4:] Conduct a Nearest Neighbor Search to identify points from $U$, nearest to the K-D Tree.
\item[Step 5:] Add these points to the minority set, form a  new labeled balanced data set $LB$ and use $LB$ to train a classifier.
\end{description}

\begin{figure}[]
\includegraphics[width=9cm]{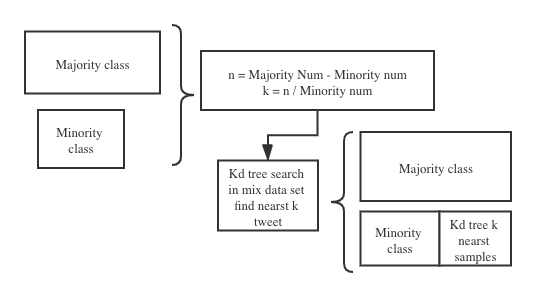} 
\centering
\caption{The K-D Tree data generation method}
\label{fig:kdtree}
\centering
\end{figure}

\begin{figure}[]
\includegraphics[width=12cm]{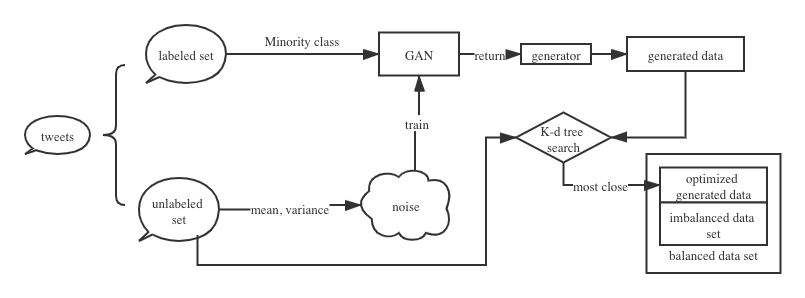}
\centering
\caption{Replace GAN's structure}
\label{fig:GAN_structure}
\centering
\end{figure}

We consider two instantiations of our framework: the direct approach and the indirect approach. The direct approach is illustrated in Figure \ref{fig:kdtree}. That approach skips step 2. In other words, it constructs a K-D Tree from the labeled minority instances present in $LI$. 
Because the minority class can contain a very small number of samples, we also propose the indirect approach which implements Step 2. The indirect approach is illustrated in Figure \ref{fig:GAN_structure}. The rationale for the indirect approach is that the minority data set may be very small and not diverse enough to help direct the search for appropriate additional instances from $U$. Generating synthetic samples which will not be included in $LB$ but which will help select actual instances from $U$, we assume, can enhance the method.

We now describe each of the steps of our algorithm in detail:
\subsubsection{Step 1: Pre-processing }
In this step, we conduct the corpus cleaning work first. Since we are working with Twitter textual data, we remove all special symbols, white spaces and emoticon icons from the content of the tweets. This helps reduce the complexity of the text content. In addition, we remove all the stop words, which forces the model to pay more attention to vocabulary with practical meaning than common terms. To minimize the external factors like the word embedding to the evaluation,
Using pre-trained checkpoints provided by the Digital Epidemiology Lab\cite{muller2020covid} as a starting checkpoint, we train a (Bidirectional Encoder Representations from Transformers) BERT model \cite{devlin2019bert}, to compute 
embedding for our tweet corpora, .

\subsubsection{Step 2: Synthetic Sample Generation}
In this step, used by the indirect approach, we generate synthetic samples by using both classical and deep-learning means. In particular, we use: SMOTE, ADASYN and a VAE-GAN.
The samples are generated according to the processes described in Section 2 for each of the approaches.
Please note that we also experimented with both a DC-GAN and a VAE, but since the results were not better than those obtained with a VAE-GAN, we decided not to include them in our graphs in order not to clutter the presentation.

\subsubsection{Step 3: K-D Tree Construction and Nearest Neighbor Search}
\paragraph{K-D Tree Construction: } In this step, we construct the tree with a recursive rule that splits the data according to the dimension/feature with highest variance. 
The dimension selected for splitting is set as the root node of the K-D Tree or subtree under consideration. This is done by finding the median for this dimension and using that value as a segmentation hyperplane, i.e., 
all the points whose value in that dimension is smaller than the median value are placed in the left child, and all the points with a greater value are placed in the right child. This procedure is followed recursively until nodes cannot be split anymore. 
%
%
%
\paragraph{Nearest Neighbor Search: } In the search query, the search starts from the root node and moves down the tree recursively. It goes to the left or right child of the node it is currently visiting depending on its relation to the node value. Once the search reaches a leaf node, the algorithm sets it as ``best current result". It then searches the other side of the parent to find out whether a better solution is available there. If so, it continues its search there, 
looking for a closer point. If such a point does not exist, it moves up the tree by one level and repeats the process. The search is completed when the root node is reached.

\subsubsection{Step 4: Balanced Data Set Formation and Classification}
To re-balance the data set, we first assume there are $n_{max}$ instances of the majority class and $n_{min}$ instances of the minority class in the data set. For each method, we augment the data using the following rules:
\begin{description}
    \item[K-D Tree:]
For each minority data $x_i$, traverse the tree composed of unlabeled data and find the $n_{aug_i}=\frac{n_{max}-n_{min}}{n_{min}}$. Add $n_{aug} = \sum_{i=1}^{n_{min}} n_{aug_i}$ to the data set after assigning them to the minority class.
\item[SMOTE, ADASYN, GAN:] Generate $n_{aug} = ({n_{max}-n_{min}})$ artificial samples, set them as minority class and add to the data set.
\end{description}
 After the data set is balanced, a logistic regression classifier is trained.

\section{Experimental Evaluation}
\subsection{Data sets} 
\paragraph{Data Set 1:} The first data set was collected for the study by \cite{boukouvalas_independent_2020} which initially randomly collected a sample of 282,201 Twitter users from Canada by using the Conditional Independence Coupling (CIC) method \cite{White2012SamplingOS}.  All tweets posted by these users between January 1, 2020 and March 13 were collected, and a random subset of 1,600 tweets was further analyzed through keyword search. A carefully curated and labeled sub data set was carved out from the random subset and includes 280 reliable and 280 unreliable Tweets and represent data set $LI$. The remaining 1,040 samples are unlabeled 
and correspond to data set $U_1$. We created a testing set $Test_1$ by  randomly selecting 40 reliable and 40 unreliable tweets from 
$LI$. 
From the rest of the labeled data, we created several ${LI_1}_n$ data sets with all the 240 reliable tweets and different numbers, $n$, of unreliable tweets, where $n$ belongs to the set $\{5,6,7,8,9,10,20,30,40,50,100,150\}$. 

\paragraph{Data Set 2:} The second data set is the COVID-19 Fake News Data set from \cite{patwa2021overview}, which includes a manually annotated data set of 10,700 social media posts and articles of real and fake news on Covid-19. We randomly selected 6,000 of them, with 3,000 true news and 3,000 fake news for $LI$. We randomly selected 100 reliable and 100 unreliable tweets from $LI$ to create our testing set, $Test_2$. To create training sets ${LI_2}_n$, we randomly selected 900 samples from the true news subset and different  numbers, $n$, from the fake news subset, where $n$ belongs to the set $\{5,6,7,8,9,10,20,30,40,50,100,150\}$. The samples that were not selected were stripped of their labels and constitute the unlabeled data set $U_2$.

\subsection{Training and Testing method} 
%
%
\paragraph{Training:} In our experiments, we trained the logistic regression classifier on the two data sets (Data Set 1 and Data Set 2), using the different data augmentation methods previously discussed to balance the training set.  In more detail, we ran the following series of experiments on both Data Sets 1 and 2. Each experiment was repeated for a minority class of size $n$ where $n$ belongs to $\{5,6,7,8,9,10,20,30,40,50,100, 150\}$. Each of the 150 generated data sets are called $LI_n$. 
\begin{itemize}
    \item Train Logistic Regression on $LI_n$. The results for this series of experiments are seen on the curve called ``Original".
    \item Train Logistic Regression on $LI_n$ augmented by: SMOTE, ADASYN, VAE-GAN.
    The SMOTE and ADASYN functions we used in our task come from the python package ``imblearn". We implemented the VAE-GAN on our own. 
    The results for this series of experiments are reported on the curves called ``SMOTE", ADASYN" and ``VAE-GAN" respectively.
    \item Train Logistic Regression on $LI_n$ augmented using the K-D Tree and Nearest Neighbor Search technique on the n instances of the minority class present in $LI_n$. We recall that that technique selects data from $U$, the unlabeled set, that most closely resembles the $n$ samples of the minority class. This the Direct implementation of our framework that skips Step 2 in the Algorithm of Section 3.  The results for this series of experiments are reported on the curve called ``K-D Tree".
    \item Train Logistic Regression on $LI_n$ augmented using the K-D Tree and nearest neighbor search technique on the $n$ instances of the minority class and their augmentations through: SMOTE, ADASYN and VAE-GAN.
 We recall that this technique selects data from $U$, the unlabeled set, that most closely resembles the n samples of the minority class and the synthetic samples generated from them using one of the generation method shown above. This is the indirect implementation of our framework that uses Step 2 in the Algorithm of Section 3. 
 The results for this series of experiments are reported on the curves called ``SMOTE-KD, ``ADASYN-KD", and ``Replace-GAN".   
\end{itemize}

\paragraph{Testing Regimen:} In total, we conducted 192 tests on 24 $LI$ data sets with different numbers of minority class samples from the 2 data sets.
The final result for each of these 192 experiments are reported based on the testing sets $Test_1$ and $Test_2$, respectively.
Since both data sets only had very few labeled data samples to use for testing, we decided to use the Bootstrap error estimation technique to evaluate the performance of our Method \cite{Japkowicz2011EvaluatingLA}\footnote{The Bootstrap technique is implemented by repeating the sampling and testing procedure previously described 100 times and using the results of these experiments to estimate the real F1, Precision and Recall values and evaluate their standard deviation.}. 
We report the F1, Precision, and Recall values of all the classifiers tested on the test set.

\section{Results} 
\begin{figure}[]
\centering\includegraphics[width=13cm]{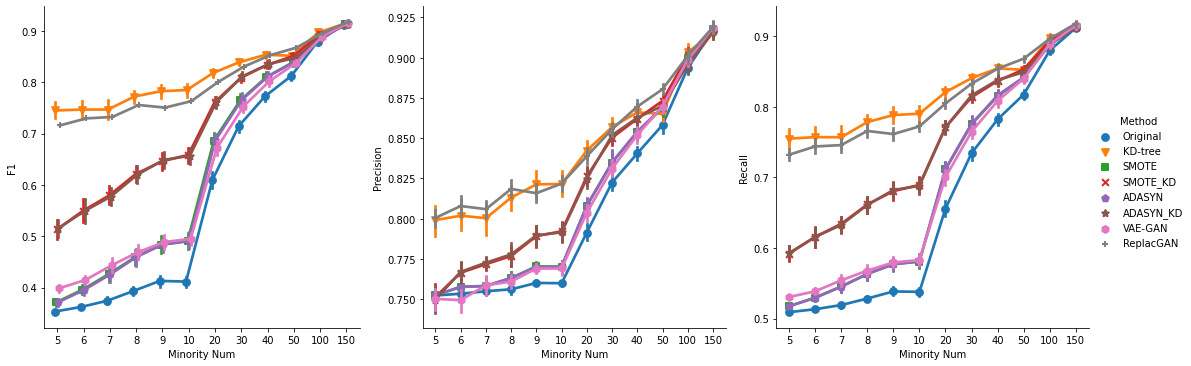} 
\centering
\caption{The F1, Precision and Recall values of Data set1}
\label{fig:result1}
\centering
\end{figure}

\begin{figure}[]
\centering\includegraphics[width=13cm]{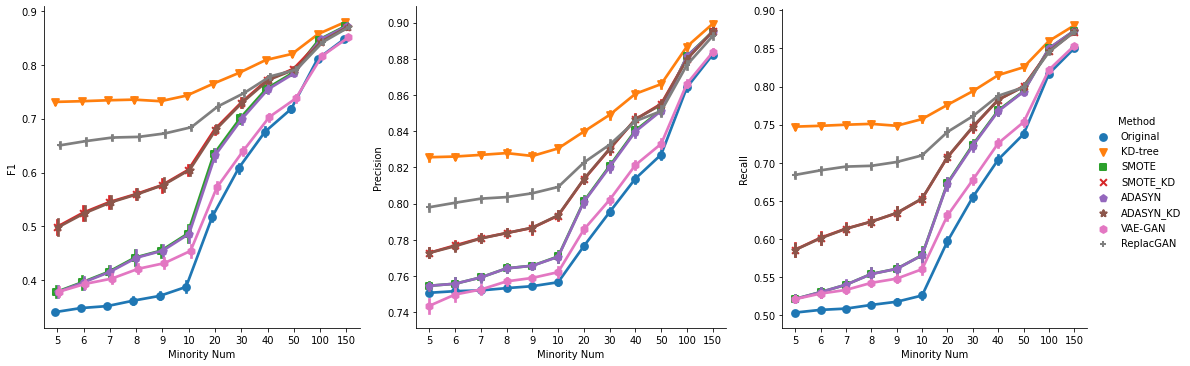}
\centering
\caption{The F1, Precision and Recall values of Data set2}
\label{fig:result2}
\centering
\end{figure}

The results of our experiments appear in Figures~\ref{fig:result1} and \ref{fig:result2}.
\bigskip
\noindent
In all the graphs, the horizontal axis represents the number of labeled minority instances used to train the Logistic Regression classifier (with or without the different kinds of minority class inflation considered in this work). To emphasize the interesting part of our work, specifically, what happens when the number of labeled minority instances is extremely small, the horizontal axis shows the results for 5, 6... 10 labeled minority samples and then jumps to 20,..50, and then to 100 and 150, where the methods produce results much closer to each other than in the very sparse case. The vertical axis represents the F1-measure, the Precision or the Recall obtained by the classifiers. 
The standard deviations at each point are indicated by a bar, visible only when it is high enough. 
The graphs show that distinct differences in the results really happen when $n$, the number of minority instances initially present, is small. As $n$ increases, the differences between the methods becomes less and less visible. We also find that the results are similar for dataset 1 and and dataset 2.

In general, we find that ``Original", where no correction for the class imbalance domain is made, obtains the worst performance. This is followed closely by the three synthetic approaches (SMOTE, ADASYN and VAE-GAN) with a slight advantage for VAE-GAN in Data Set 1 and  a slight advantage for SMOTE and ADASYN (which show identical performance in all experiments) in Data Set 2. As shown in all graphs, the advantage gained by these synthetic resampling methods is modest. Next, in terms of performance, come the three indirect methods of our framework, SMOTE-KD, ADASYN-KD, ReplaceGAN. We recall that these are the methods that generate synthetic data but do not use them directly. Instead, they are used to identify appropriate unlabeled samples to add to the minority class. The results show that these approaches obtain noticeably higher F1, Precision and Recall results, with a distinct advantage for ReplaceGAN. This is true for both data sets, and suggests that the addition of real data through our semi-supervised scheme rather than synthetically generated data is a superior proposition. 
Finally, the results show, that in both domains, using the Direct implementation of our framework yields a better performance than the ReplaceGAN strategy. That difference is slight in Data Set 1, where the standard decision bars indicate that ReplaceGAN, while slightly less accurate, is more stable than K-D Tree, but it is unmistakable in Data Set 2 where ReplaceGAN is noticeably less accurate than K-D Tree. This suggests that our hypothesis regarding the advantage that a greater diversity to start off our unlabeled data set search for minority sample candidates did not pan out and the indirect implementation of our framework is less desirable than its Direct implementation.

While we commented on the results qualitatively, some quantitative remarks are in order. For $n = 5$, the difference between the F1 value of the methods is remarkable. In both domains, the results obtained by ``Original" are below .4. They get near or reach .4 with the synthetic resampling methods. The semi-supervised indirect methods SMOTE-KD and ADASYN-KD yield F1 measures around .5 in both domains while ReplaceGAN reaches an F1 measure above .7 for Domain 1 and between .6 and .7 for Domain 2. Finally, the semi-supervised Direct K-D Tree method obtains an F1-measure well over .7 in each domain. Until $n = 10$, a similar trend is observed. By $n = 50$, however, the fluctuation of all the methods lies in a much smaller interval since the F1 measures are all between slightly over .7 and slightly aver .8 for Domain 1; and between slightly below .7 and slightly above .8 for Domain 2.For higher values of $n$ all methods become equivalent. This shows that the impact of our framework is much more significant in extreme class imbalance cases than in more moderate ones.

\begin{figure}[]
\includegraphics[width=6cm]{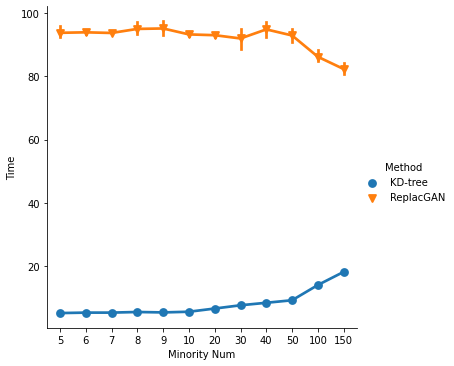}
\includegraphics[width=6cm]{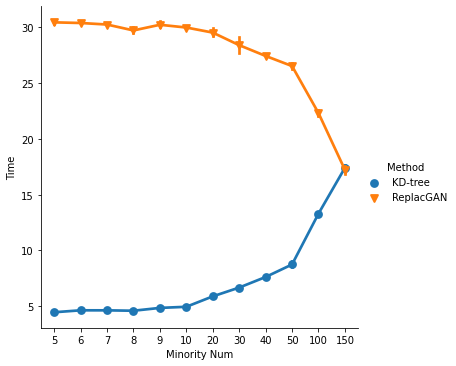}
\centering
\caption{The running time of K-D Tree and ReplaceGAN on Data set 2 with 3,000 reliable instances (left) and 300 reliable instances (right).}
\label{fig:runtime}
\centering
\end{figure}

In terms of run time, we tested the K-D tree and ReplaceGAN's running time with data set 2. The results are shown in figure \ref{fig:runtime} for 3,000 and 300 reliable samples. We use Python's built in time function to calculate the running time for each epoch. We run each method with different number of minority samples and report the results as an average of 50 times. Time is measured in seconds.
As expected, we found that the K-D Tree has a much lower running time than ReplaceGAN. This is because the K-D Tree search method needs this time to conduct many fewer root-to-leaf search queries than it does in the case of the ReplaceGAN strategy due to the smaller number of instances present. Interestingly, however, this discrepency is less important and eventually vanishes in smaller data sets and with larger amounts of labeled minority samples, where the stability of ReplaceGAN is also greater. 


\section{Discussion}

In this paper, we presented a semi-supervised framework for identifying appropriate unlabeled samples to inflate the minority class and create a more balanced data set to learn from. The framework was designed specifically for non-continuous domains such as text, and tested on two misinformation/Fake news detection data sets where it obtained remarkable results, especially in cases of extreme class imbalance. Two categories of approaches of the framework were tested: the direct and indirect approach. The direct approach (K-D Tree) performed better than the indirect approach using a GAN (ReplaceGAN) but was not as stable in the smaller dataset (dataset 1). The direct approach is also more efficient than the indirect one, but the disparity is less noticeable in smaller data sets. 
The results obtained with our framework were significantly better than those obtained by methods that augment the data by synthetic generation, thus supporting the assumption that synthetic generation in non-continuous domains such as Text is not particularly useful and that semi-supervised methods such as ours fare much better.



In the future, we propose to investigate the utility of the ReplaceGAN indirect approach more carefully. We will also extend our framework to different domains (e.g., the genetic domain and images) including continuous and discrete ones where an unlabeled data set exists, and test other classifiers on our resulting augmented data sets. This will allow us to test whether the advantage we noticed in text data and with logistic regression carries over to other types of domains and classifiers as well. We will also apply our method to less extremely imbalanced data sets but use it in a finer grained manner, using a decomposition of the classes into sub-classes prior to re-sampling from the unlabeled set. This, we believe, will allow us to counter the kind of biases and unfairness introduced by incomplete data sets. More generally, we will also attempt to use our framework in the context of a data labeling tool having only a few seed labels to start from.

\section*{Acknowledgement}
Computing resources used for this work were provided by the American University Zorro High Performance Computing System. \\
Pre-trained Bert model from Digital Epidemiology Lab EPFL.\\

\bibliographystyle{acm} 
\bibliography{ref}

\end{document}